\RequirePackage{fix-cm}
\documentclass[smallextended]{svjour3}       
\smartqed  
\usepackage{graphicx}
\usepackage{graphicx}
\usepackage{siunitx}
\usepackage{multirow}
\usepackage[nolist]{acronym}
\usepackage{subfig}
\usepackage{makecell}
\usepackage{amsmath}
\usepackage{mathrsfs}

\usepackage{xcolor}
\usepackage{siunitx}
\usepackage[dash,dot]{dashundergaps}

\newcommand{\figref}[1]{Fig.~\ref{#1}}
\newcommand{\secref}[1]{Section~\ref{#1}}
\newcommand{\tabref}[1]{Table~\ref{#1}}

\usepackage{contour}
\usepackage[normalem]{ulem}

\contourlength{0.8pt}

\renewcommand{\underline}[1]{%
	\uline{\phantom{#1}}%
	\llap{\contour{white}{#1}}%
}

\begin{document}
\begin{acronym}[]
	\acro{AAAR}{Abdominal Aortic Aneurysm Repair}
	\acro{AAA}{Abdominal Aortic Aneurysm}
	\acro{AC}{Alternating Current}
	\acro{ANN}{Artificial Neural Network}
	\acro{BCE}{binary-cross-entropy}
	\acro{CMM}{Coordinate Measuring Machine}
	\acro{DC}{Direct Current}
	\acro{DoF}{Degree of Freedom}
	\acroplural{DoF}{Degrees of Freedom}
	\acro{DNN}{Deep Neural Network}
	\acroplural{DNNs}{Deep Neural Networks}
	\acro{EKF}{Extended Kalman Filter}
	\acro{EMT}{ElectroMagnetic Tracking}
	\acro{EMTS}{Electromagnetic Tracking System}
	\acro{EVAR}{Endovascular Aneurysm Repair}
	\acro{FFNN}{Feed Forward Neural Network}
	\acro{FPGA}{Field Programmable Gate Array}
	\acro{GAN}{Generative Adversarial Network}
	\acro{HP-MIS}{High Precision Minimally Invasive Surgery}
	\acro{LOS}{Line of Sight}
	\acro{MC}{Monte Carlo}
	\acro{MIS}{Minimally Invasive Surgery}
	\acro{MSE}{Mean-Squared-Error}
	\acro{NDI}{Northern Digital Inc.}
	\acro{OR}{Operating Room}
	\acro{ROI}{Region of Interest}
	\acro{SLAM}{Simultaneous Localization and Mapping}
\end{acronym}

\title{CycleGAN for Interpretable Online EMT Compensation}

\titlerunning{CycleGAN for Interpretable Online EMT Compensation}
%
\author{Henry Krumb \and
        Dhritimaan Das \and
        Romol Chadda \and
        Anirban Mukhopadhyay
}
%
%
\institute{
    H. Krumb \and R. Chadda \and A. Mukhopadhyay \at Technische Universit\"at Darmstadt, Germany \and
    D. Das \at Indian Institute of Technology (IIT-BHU), Varanasi, India
}

\date{Received: date / Accepted: date}

%
\maketitle
\begin{abstract}
\noindent\textbf{\\* Purpose:}
\ac{EMT} can partially replace X-ray guidance in minimally invasive procedures, reducing radiation in the OR.
However, in this hybrid setting, \ac{EMT} is disturbed by metallic distortion caused by the X-ray device.
We plan to make hybrid navigation clinical reality to reduce radiation exposure for patients and surgeons, by compensating \ac{EMT} error.


\noindent\textbf{\\* Methods:}
Our online compensation strategy exploits cycle-consistent generative adversarial neural networks (CycleGAN).
3D positions are translated from various bedside environments to their bench equivalents.
Domain-translated points are fine-tuned to reduce error in the bench domain.
We evaluate our compensation approach in a phantom experiment.

\noindent\textbf{\\* Results:}
Since the domain-translation approach maps distorted points to their lab equivalents, predictions are consistent among different C-arm environments.
Error is successfully reduced in all evaluation environments.
Our qualitative phantom experiment demonstrates that our approach generalizes well to an unseen C-arm environment.

\noindent\textbf{\\* Conclusion:}
Adversarial, cycle-consistent training is an explicable, consistent and thus interpretable approach for online error compensation.
Qualitative assessment of \ac{EMT} error compensation gives a glimpse to the potential of our method for rotational error compensation.

\end{abstract}

\acresetall

\section{Introduction}
In minimally invasive surgery, \ac{EMT} has the potential to partially replace continuous X-ray navigation \cite{krumb2020leveraging}, reducing the radiation exposure to both patients and surgeons.
Such procedures are traditionally performed under X-ray only (for example laparoscopy \cite{aoki2020laparoscopic}, \ac{EVAR} \cite{dijkstra2011intraoperative}).
Our vision is to enable hybrid navigation in the clinical setting, where EMT can replace X-ray as the primary continuous tracker, and X-ray snapshots are only acquired intermittently.
However, in current practice \ac{EMT} is susceptible to metallic distortion caused by the C-arm, such that the surgeon can put little trust in \ac{EMT} in between snapshots.


Traditional error-compensating algorithms to increase trust in EMT are offline in nature, resulting in tedious calibration and impractical clinical translation.
We thus advocate learning-based online error compensation where a general purpose learning model is trained \textit{only once}.
Online compensation of \ac{EMT} error can be realized by implementing data-driven models, which generalize among data from different environments.
In previous works, we investigated learning-based online EMT compensation by employing a series of increasingly complex learning models (polynomial regression \cite{kuegler2019high}, \acp{ANN} \cite{krumb2020leveraging}).

Despite giving increasingly better results, the interpretation of the failure modes for these complex models is a growing concern in learning literature \cite{tjoa2020survey}.
Our major focus is to develop an \textit{interpretable} error compensation technique, which imposes two more constraints beyond mere error reduction:
\textit{explicability} and \textit{consistency}.
Firstly, predictions need to be \textit{explicable}; that is we need to know \textit{why} a certain point from C-arm domain is mapped to a specific compensated point.
Otherwise, compensation results could be arbitrary and still fulfill topological constraints, giving a false sense of reliability.

Our second constraint is \textit{consistency} among distorted environments.
Consistency is important for \textit{online} error compensation, where training data stems from environments with varying distortion characteristics.
If output is inconsistent among input environments, changing distortion characteristics of the electromagnetic field (e.g. by moving the C-arm in the course of a surgical procedure) affect the coordinate frame of compensated points.
Such changes jeopardize the X-ray-to-EMT registration, rendering our envisioned hybrid setting infeasible.

To impose these two constraints, we approach the problem as domain adaptation with cyclic-consistent \acp{GAN} \cite{goodfellow2014generative,zhu2017unpaired}.
The main purpose of our approach is to learn a mapping from points of domain $C$, which is the domain of C-arm-distorted points from environments similar to bedside, to domain $L$, which is the domain of laboratory bench settings.
Our approach is both explicable and consistent by design, and thus more interpretable than our \ac{ANN} \cite{krumb2020leveraging} approach.
While domain adaptation using \acp{GAN} is quite popular in the medical imaging setting \cite{kazeminia2020GANs}, it has never been studied before in the context of \ac{EMT} error compensation.

\section{Related Work}
Offline compensation methods, such as interpolation schemes or polynomial fits, are extensively described in the comprehensive reviews by Kindratenko et al. \cite{kindratenko2000survey} and Franz et al. \cite{franz2014electromagnetic}.
For the sake of brevity, we refer the reader to these papers.

An \textit{online} compensation framework is proposed by Sadjadi et al. in 2016 \cite{sadjadi2016simultaneous}.
Their approach combines \ac{SLAM} and an extended Kalman filter to rectify \ac{EMT} measurements by $67\%$.
Although this approach is promising, it presumes a multi-sensor-setup, which is hardly feasible in minimally invasive applications with narrow access canals.

\begin{figure}
    \centering
    \includegraphics[width=0.8\linewidth]{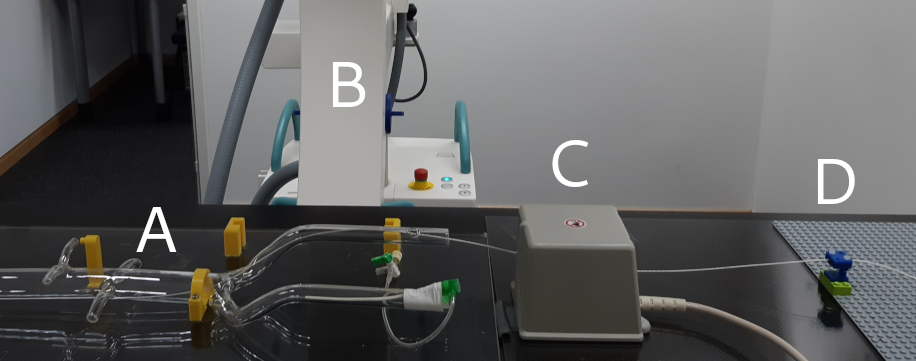}
    \caption{Hybrid navigation experiment with Aortic phantom. A) phantom with EMT sensor inside, B) C-arm, C) EMT field generator, D) sensor cable fixture on Lego board}
    \label{fig:c-arm-aorta}
\end{figure}


Neural networks for error compensation are employed by Kindratenko et al. in 2005 \cite{kindratenko2005neural}.
The authors use two-layer feed forward networks to compensate \ac{EMT} error in an offline setup.
In our previous work \cite{krumb2020leveraging}, we show that \acp{ANN} not only work for offline compensation, but are capable of generalizing among different distortion scenarios (online) as well.
The \acp{ANN} learn a non-linear transform which reduces distance error between measured points.
Effectively, this method alters the point topology.

In this paper, compensation is performed by translating individual points from C-arm to laboratory domain using generative adversarial neural networks (\acp{GAN}).
This adversarial domain adaptation approach is based on the CycleGAN \cite{zhu2017unpaired} architecture, which is originally developed for the application of image-to-image translation.
While the topology-optimizing \ac{ANN} fails silently (\figref{fig:aorta-eval}), predictions made by our proposed online approach are interpretable by design.

\section{Materials}
Our data-driven compensation approach uses measurements we acquired during our previous work \cite{krumb2020leveraging}, and were collected using an Ascension 3D Guidance trakSTAR system (stated accuracy of \SI{1.4}{\milli\meter}), a 180-type sensor (6 DOF, \SI{1.8}{\milli\meter} in diameter) and mid-range field generator.
Custom C++ software is used for data collection with the trakSTAR system.
We use a calibrated Lego board in different elevations to collect data in three positional \acp{DoF}, as proposed by us earlier \cite{kuegler2019high}.
For each measuring point, 500 samples are collected and averaged to reduce random noise.
Displacement distances between points on the Lego board are then used as ground truth to calculate displacement error.
This metric, based on relative displacements, eliminates the need for an additional measurement standard (e.g. a ruler).
Positional error of displacements we use for training, validation and evaluation are listed in \tabref{tab:scenarios}.

Similarly to other \ac{EMT} systems, the trakSTAR provides a quality estimate with each measuring point, indicating the amount of metallic distortion. Each measuring point is constituted by $(x,y,z,q,\phi_{x},\phi_{y},\phi_{z})$, where $q$ denotes the system's quality estimate and $\phi_x$, $\phi_y$ and $\phi_z$ denote rotation around x, y and z axes.

In our phantom study (\secref{sec:phantomstudy}), we place the sensor at different positions inside an acrylic glass phantom, resembling a human aorta in life size.
A 3D printed Lego adapter holds the sensor cable (\figref{fig:c-arm-aorta}) in place.
Measurements for the phantom study and other bedside measurements are performed in vicinity of a Ziehm Vision RFD C-arm device.

Our learning models are implemented in Python using the PyTorch \cite{pytorch} framework.

\begin{table}
	\centering
	\begin{tabular}{c|c|c|c|c}
		& scenario & \thead{\#points} & \thead{displacement\\RMSE [mm]} & \thead{displacement\\ std. dev. [mm]} \\ 
		\hline
		\multirow{4}{*}{training} & \shortstack[l]{\\laboratory\textsuperscript{1}} & \shortstack[l]{\\60} & \shortstack[l]{\\ \textbf{0.367} } & \shortstack[l]{\\ \textbf{0.202} } \\
		& c-arm \SI{8}{\centi\meter} & 60 & 1.292 & 1.264 \\
		& c-arm \SI{11}{\centi\meter} & 60 & 1.064 & 0.917 \\
		& c-arm\textsuperscript{2} \SI{50}{\centi\meter}\ & 60 & 0.639 & 0.309 \\
		\hline
		\multirow{1}{*}{validation} & c-arm \SI{10}{\centi\meter} & 60 & 1.101 & 0.989 \\
		& c-arm\textsuperscript{3} \SI{30}{\centi\meter} & 60 & 0.743 & 0.372 \\
		\hline
		\multirow{3}{*}{evaluation} & c-arm \SI{7}{\centi\meter} & 60 & 1.386 & 1.389 \\
		& c-arm \SI{9}{\centi\meter} & 60 & 1.192 & 1.139 \\
		& c-arm \SI{12}{\centi\meter} & 60 & 1.025 & 0.833 \\
	\end{tabular}
	\caption{%
		Datasets collected in varying distances to c-arm and in a laboratory setup.
		Number of displacements, RMSE and standard deviation are noted for each dataset.
		Distances to c-arm are measured from x-ray source to base board center.
		\textsuperscript{1}: only 2 of 3 z-layers used for training,
		c-arm\textsuperscript{2}:  gantry rotated at $60^\circ$, c-arm\textsuperscript{3}: gantry rotated at $30^\circ$
	}
	\label{tab:scenarios}
\end{table}

\section{Methods}
\label{sec:methods}


To fulfill the explicability constraints we identified for interpretable online error compensation, we modify a domain adaptation approach that is originally used for image translation tasks.
Instead of translating between two image domains (e.g. photorealistic vs. abstract), our goal is to translate \ac{EMT} measurements from the bedside (high error) to the bench (low error) domain.
Since the objective of this translation task is intuitive, we expect this compensation to be \textit{explicable}.

In the following, we detail the modified domain adaptation architecture, the protocol and parameters we use for training.
Subsequently, we describe how error in laboratory domain is compensated by a post-processing step, which uses a simple linear regression model.

\subsection{Domain Adaptation by Adversarial Training}
\begin{figure}
    \centering
    \includegraphics{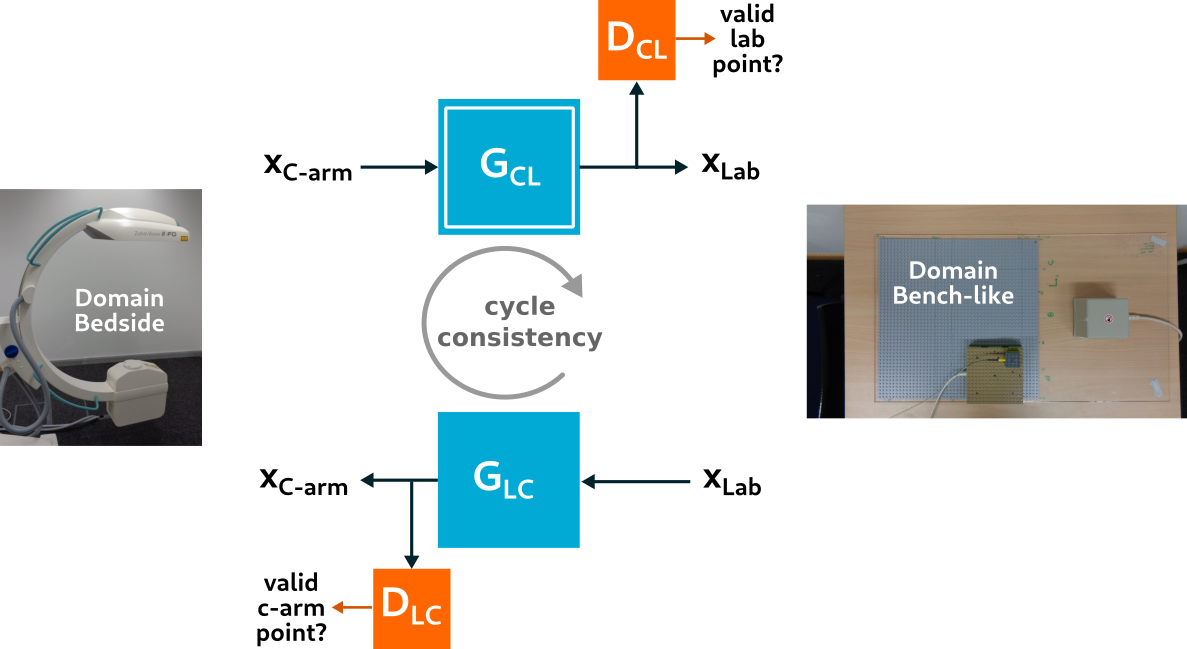}
    \caption{Cycle-consistent GAN architecture for unsupervised domain adaptation. $G_{CL}$ translates points from C-arm to lab domain and, once trained, is used for compensation.}
    \label{fig:cycleGAN}
\end{figure}

We employ cycle-consistent adversarial training for interpretable \ac{EMT} compensation.
Similar to the work of Zhu and Park et al. \cite{zhu2017unpaired}, we make use of two different \ac{GAN} models, one for each direction of domain translation (C-arm to lab, lab to C-arm).
As illustrated in Figure~\ref{fig:cycleGAN}, the training process connects both \acp{GAN} to achieve cyclic consistency.
Since input data from laboratory and C-arm (Table \ref{tab:scenarios}) are unpaired and translation is thus under-constrained, adversarial training benefits from this additional cycle-consistency constraint.

Each of the two \acp{GAN} consists of a generator network and a discriminator neural network.
The generator receives an input point and generates a domain-adapted point, whereas the discriminator judges whether the generated point stems from the target domain.
The generator's objective is to trick the discriminator by generating points close to its target domain, given an input point from the original domain.
For instance, $G_{CL}$ takes a point from C-arm environment and tries to generate a corresponding laboratory point, and $D_{CL}$ learns to judge whether this point actually is a \textit{valid} laboratory point.
Ideally, both parties in this adversarial two-player game achieve the Nash equilibrium \cite{goodfellow2014generative,nash1951non}.

Our two generator models (and discriminator models respectively) share an identical structure.
Generators receive a $(x, y, z, q, \phi_x, \phi_y, \phi_z)$ point (normalized to $[0,1]$) as input and produce a vector $(\hat{z}, \hat{q}, \hat{\phi_x}, \hat{\phi_y}, \hat{\phi_z})$, where $\hat{z}$, $\hat{q}$, $\hat{\phi_x}$, $\hat{\phi_y}$, $\hat{\phi_z}$ denote domain-translated values for $z$, quality and orientation.


To ensure that points from C-arm domain are not translated to \textit{arbitrary} points in the lab domain, the generators do not alter the $x$ and $y$ positional components.
We focus only on compensating the $z$ component, since of all positional components, it is the most susceptible to error (for the trakSTAR system).
Error in the $x$-$y$-plane is compensated in a fine tuning step further described in section \secref{sec:finetuning}.

\subsection{Training Protocol}
Our training protocol is similar to that of CycleGAN, but includes additional loss terms tailored to the problem of \ac{EMT} error compensation.
In particular, we compute the generator loss as weighted sum of individual penalties, which are described in the following:

\paragraph{Adversarial Loss}
$L_{adv}$ is a \ac{BCE} loss term, which reflects how well each generator can fool its corresponding discriminator.
It is computed as

\begin{equation*}
    L_{adv} = BCE\big(D_{CL}(G_{CL}(x_{C})), l_{valid}\big) + BCE\big(D_{LC}(G_{LC}(x_{L})), l_{valid}\big)
\end{equation*}

\noindent Where $l_{valid}$ denotes the discriminator label we assign to valid points, that is the label the discriminator would assign to points that truly stem from the target domain.

\paragraph{Cycle Loss}
As described in the original CycleGAN paper \cite{zhu2017unpaired}, we enforce cycle-consistency by adding a loss term $L_{cycle}$:

\begin{align*}
    L_{recov,L} &= |G_{CL}(G_{LC}(x_{L})) - x_{L}|\\
    L_{recov,C} &= |G_{LC}(G_{CL}(x_{C})) - x_{C}|\\
    L_{cycle} &= L_{recov,L} + L_{recov,C}
\end{align*}

\noindent $L_{recov,L}$ indicates how well an input point is recovered after translating it from domain $L$ to domain $C$ and back to $L$ ($C \rightarrow L \rightarrow C$ in $L_{recov,C}$ respectively).

\paragraph{Compensation Loss}
In addition to the CycleGAN losses, we also penalize distance error only in the laboratory domain (we cannot enforce error to be low in generated C-arm samples) as a means of regularization.
Compensation loss is computed as $L_{comp} = MSE(d_{G_{CL}}, d_{true})$
where $d_{G_{CL}}$ and $d_{true}$ are distances between pairs of points, which stem from $G_{CL}$ or ground truth data respectively.

The total generator loss is $L_{total} = \lambda_{adv} \cdot L_{adv} + \lambda_{cycle} \cdot L_{cycle} + q_{CL}^2$ with coefficients $\lambda_{adv}=0.5$, $\lambda_{cycle}=10$ and $\lambda_{comp}=10^{-5}$, which we determine empirically.
By penalizing high values of $q_{CL}$ we enforce generated laboratory points with low distortion estimates and thus lower error.

\subsection{Prediction Uncertainty}
Although neural networks are known to generalize well to unseen data, predictions made under a lack of knowledge are uncertain.
Fortunately, this uncertainty can be approximated by training the same architecture multiple times, but initialized with different random seeds (deep ensembling \cite{lakshminarayanan2017simple}).
Computing the standard deviation among the resulting predictions yields an approximation of model-inherent (epistemic) uncertainty, and averaging the predictions is expected to yield better prediction accuracy, since we combine the knowledge of multiple models.
We choose to sequentially train $10$ different initializations in an ensemble as a compromise between training time and accuracy.

\subsection{Network and Training Parameters}
Our discriminator models have three layers each, with 16 nodes per layer.
All layers use LeakyReLU activations with a leak of $0.2$, except for the last layer, which is Sigmoid-activated.
The discriminators are trained with soft labels (uniform distribution of $0.0..0.2$ and $0.8..1.0$ respectively) \cite{szegedy2016rethinking}.

The generators also have four layers each, with 16 nodes per layer.
Similar to the discriminators, the generator's layers use LeakyReLU activations, but with a leak of $0.01$.
The last layer uses linear activation.

Generators and discriminators are optimized under the use of Adam optimizer \cite{kingma2014adam} both with a learning rate of $0.0005$, which linearly decays to $0$ after $100$ epochs.
Whereas the generators share a common optimizer, the discriminators are both trained by individual optimizers.
During training, we use mini batches with a batch size of $16$.
The whole training for each model in the ensemble lasts 200 epochs.
This training protocol is similar to that of the original CycleGAN implementation \cite{zhu2017unpaired}.

\subsection{Fine-Tuning}
\label{sec:finetuning}

As the cycle-consistent \ac{GAN} model does not affect the x and y components of input points, there still exists positional error on the x-y-plane.
Assuming that the points compensated by $G_{CL}$ always lie in the laboratory domain, we can apply a compensation approach tailored to the laboratory domain.
For the sake of simplicity, we choose to fit a linear regression model that compensates distance error similarly to our previous work \cite{kuegler2019high}, using input features $(x, y, z, q)$.


\section{Results}
We first perform a bedside phantom evaluation of online error compensation.
Afterwards, we compare our domain adaptation approach to the \ac{ANN} we have previously proposed \cite{krumb2020leveraging}.

\begin{figure}[h]
    \centering
    \includegraphics[width=\linewidth]{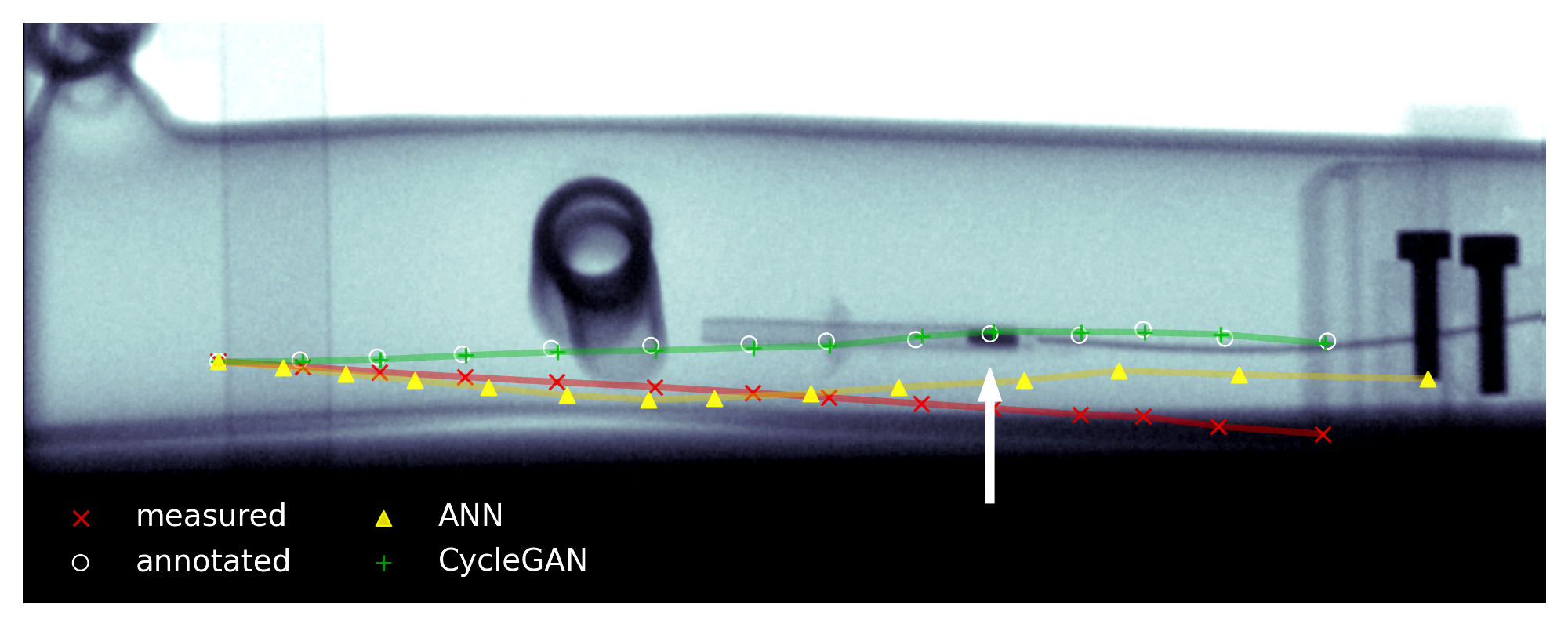}
    \caption{X-ray snapshot of aortic phantom with sensor inside. (Please find an animated video in supplementary material.) Arrow points to EMT sensor tip (dark rectangle). White circles are manually annotated sensor center points. Red, yellow and green overlays show uncompensated, ANN-compensated and domain-adapted EMT points respectively.}
    \label{fig:aorta-eval}
\end{figure}

\subsection{Bedside Evaluation on Aortic Phantom}
\label{sec:phantomstudy}

To assess the quality of our online compensation method in a realistic hybrid bedside setting, we combine EMT and X-ray imaging in a pilot phantom study.
Figure \ref{fig:c-arm-aorta} shows the measuring setup used in this experiment, in which the EMT sensor is placed inside an aortic phantom which is positioned close to the C-arm.
Since the C-arm gantry is rotated to $90^\circ$, this setup is different from anything the CycleGAN has seen during training.
Using a custom 3D printed Lego fixture, we pull out the sensor in 13 steps of \SI{8}{\milli\meter}.
For each individual step, an X-ray snapshot is created in the median plane, which corresponds to the x-z-plane of the \ac{EMT} coordinate system.

The EMT sensor can be clearly distinguished from its background on the X-ray (\figref{fig:aorta-eval}).
We could therefore annotate the EMT sensor's center points in all 13 snapshots by hand.
Points measured by \ac{EMT} are scaled to pixel dimensions, and translated to match the annotated point (leftmost in \figref{fig:aorta-eval}).
Rotation angle of the whole trajectory is estimated from compensated $\phi_y$ at the first measuring point.
We use the same transform for all three sets of points (uncompensated, \ac{ANN} and CycleGAN) and omit the fine tuning step, to allow for better comparison between uncompensated and compensated trajectories.
\figref{fig:aorta-eval} shows that our compensated points are close to the annotations, indicating that our domain-adaptation model generalizes well to the unseen environment.

\begin{figure}
    \centering
    \includegraphics[width=0.8\linewidth]{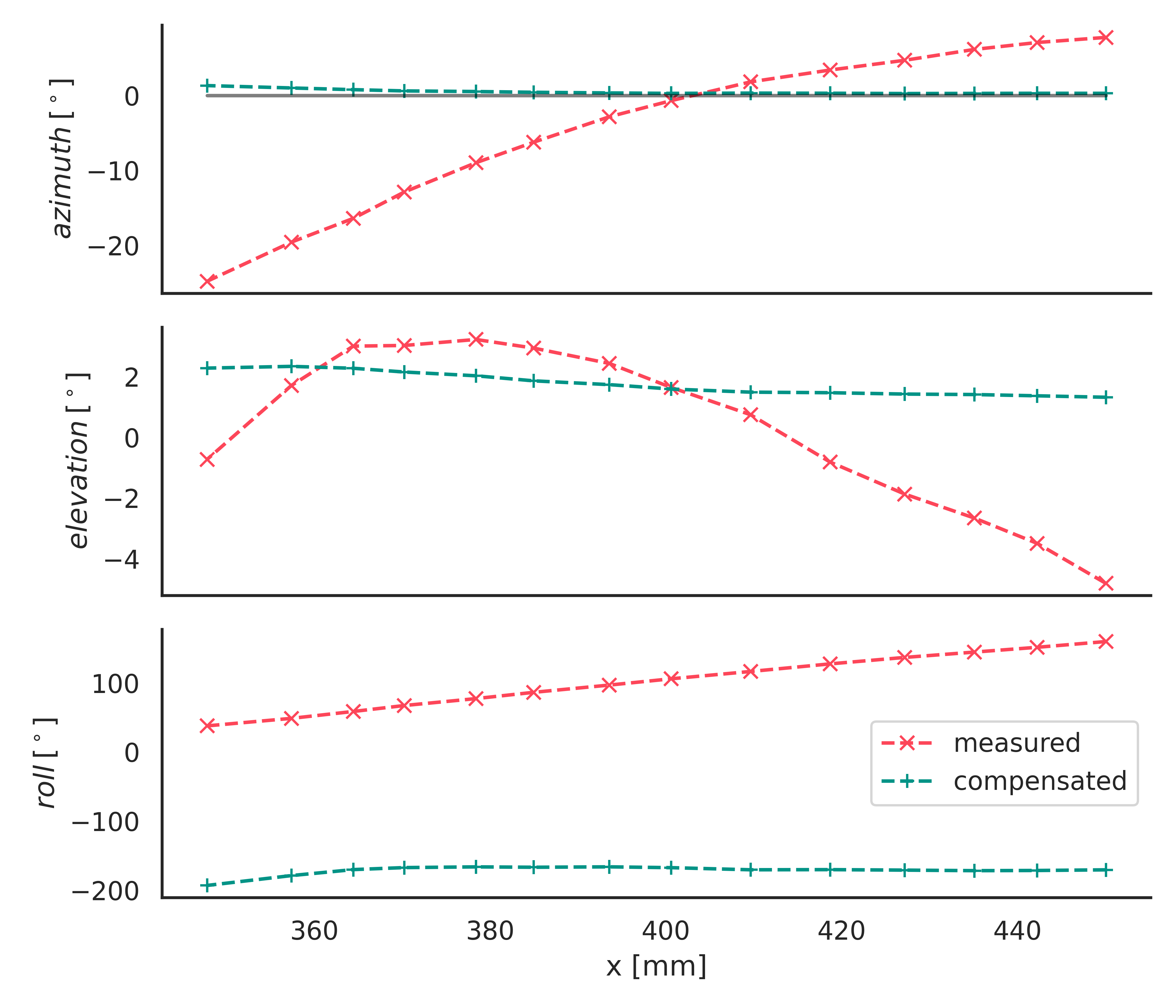}
    \caption{Measured and CycleGAN-compensated rotation angle (azimuth, elevation, roll) over x position, during sensor retraction in aorta phantom study.}
    \label{fig:phantom_rotation}
\end{figure}

Since our compensations alters rotation angles $\phi_x$, $\phi_y$ and $\phi_z$ as well, the phantom study in the C-arm environment also allows for a pilot qualitative assessment of rotational error compensation.
Although we cannot directly measure the actual orientation of the sensor inside the phantom, we can make three assumptions:

\begin{enumerate}
    \item The sensor tip is heading in positive direction of the tracker's x-axis throughout the whole experiment. We expect azimuth angle to be constant and close to zero.
    \item Elevation is almost constant and near zero. Since the aortic phantom is slightly curved, elevation should decrease with higher x.
    \item Roll angle is hard to determine absolutely (only relatively). However, we know that it does not change substantially during the experiment, as the cable is not twisted and is rigidly attached to the Lego block.
\end{enumerate}

Figure \ref{fig:phantom_rotation} shows orientations over positions on the x-axis, which are measured in our phantom study.
It illustrates that all three assumptions hold for compensated values:
1) Azimuth is constant and close to zero degrees (gray line), 2) elevation is almost constant at about $2^\circ$ and 3) roll is nearly constant.
Contrary to this, raw measurements violate all three assumptions.

\subsection{Quantitative Comparison}

\begin{figure}[h]
    \centering
    \includegraphics[width=1.0\linewidth]{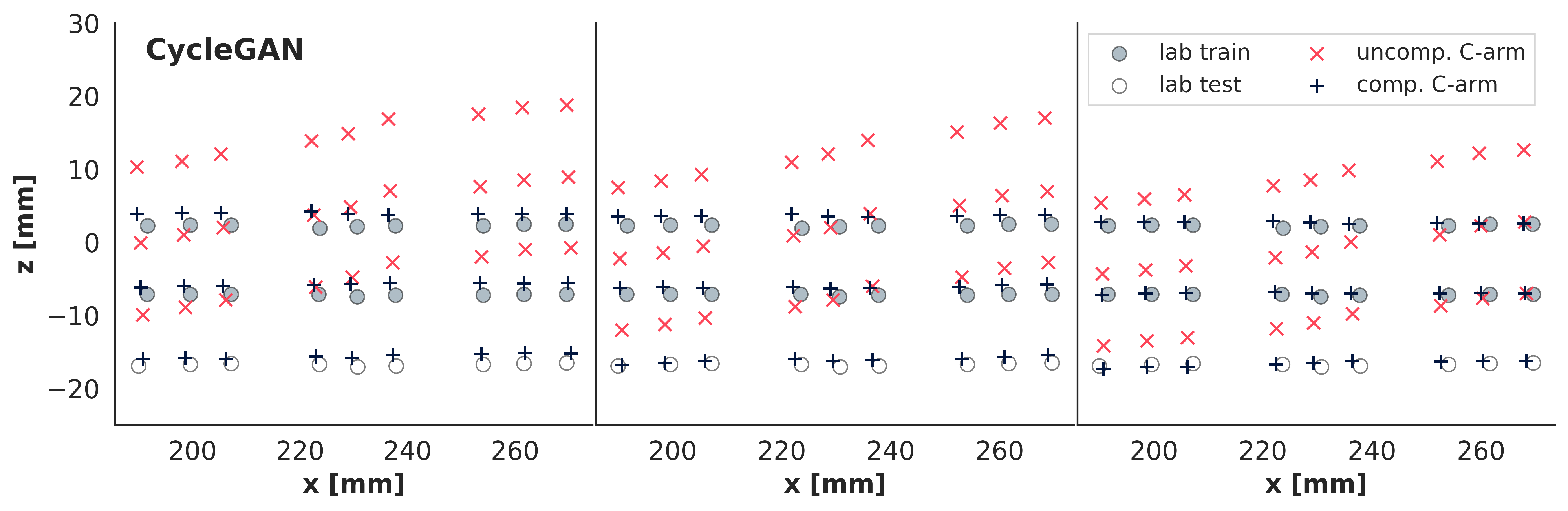}
    \label{fig:eval_cyclegan}

    \includegraphics[width=1.0\linewidth]{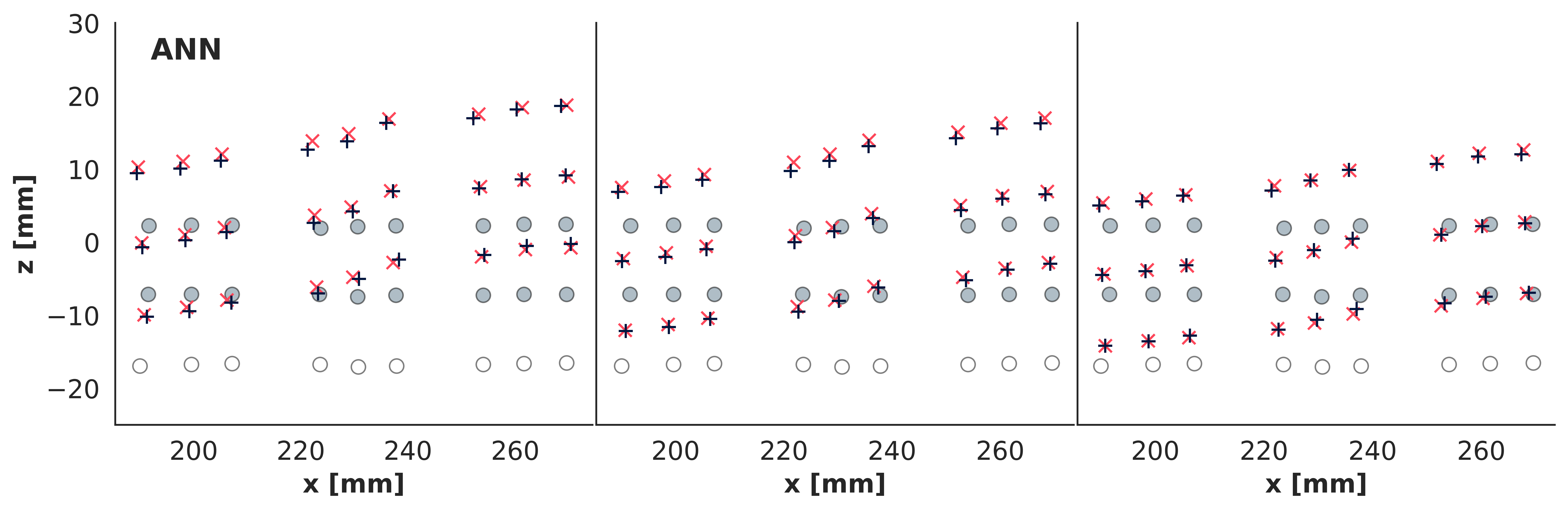}
    
    \caption{CycleGAN-adapted (top) and ANN-compensated (bottom) measuring points from C-arm at 7 cm (left), 9 cm (center), 12 cm (right), compared to corresponding lab points. White lab points are used for testing only.}
    \label{fig:eval_ffnn}
\end{figure}

In Figure \ref{fig:eval_ffnn}, we see that the domain adaptation approach is more consistent among distorted environments and yields results close to laboratory points.
Even C-arm points without a corresponding lab point in the training set are matched to their corresponding point in the laboratory domain, indicating that our method generalizes well.

On the other hand, tolopogy-based compensation by our previously proposed \ac{ANN} \cite{krumb2020leveraging} does not yield consistent output among input environments.
Actually, \ac{ANN} compensated points are still close to the input points on the z-axis and several millimeters away from corresponding lab points.

\subsection{Ablation Experiment}
Adversarial training with cycle-consistency is beneficial for compensation performance, compared to training a single \ac{GAN} translating C-arm to laboratory points.
Vanilla \ac{GAN} without cycle loss worsens overall compensation performance, as we show in \tabref{tab:results}.

Our fine-tuning step brings an additional boost in accuracy to the proposed CycleGAN setup (\tabref{tab:results}).
However, this step comes with the cost of introducing another source of predictive uncertainty.

\begin{table}[h]
    \centering
    \begin{tabular}{l | c | c | c | c}
        Method & Dataset & RMSE [mm] $\downarrow$ & $\sigma_{error}$ [mm] $\downarrow$ & $\sigma_{pred}$ [mm] $\downarrow$ \\
        \hline
                 & \SI{7}{\centi\meter} & 1.295 & 1.264 &  \\
        CycleGAN & \SI{9}{\centi\meter} & 1.090 & 0.949 & \textbf{0.370} \\
                 & \SI{12}{\centi\meter} & 1.007 & 0.722 &  \\
        \hline
        CycleGAN  & \SI{7}{\centi\meter} & \textbf{1.100} & \textbf{1.117} & \\
         + Fine   & \SI{9}{\centi\meter} & \textbf{0.811} & \textbf{0.759} & $0.768^1$ \\
           Tuning & \SI{12}{\centi\meter} & \textbf{0.622} & \textbf{0.530} & \\
        \hline
                 & \SI{7}{\centi\meter} & 1.400 & 2.744 & \\
        GAN      & \SI{9}{\centi\meter} & 1.176 & 1.890 & 0.654 \\
                 & \SI{12}{\centi\meter} & 1.030 & 1.256 & \\
        \hline
         & \SI{7}{\centi\meter} & \textbf{0.815} & \textbf{0.825} & \\
        ANN \cite{krumb2020leveraging} & \SI{9}{\centi\meter} & \textbf{0.325} & \textbf{0.628} & 0.751 \\
             & \SI{12}{\centi\meter} & \textbf{0.445} & \textbf{0.406} & \\
        \hline
        ANN & \SI{7}{\centi\meter} & 1.391 & 2.844 & \\
        xy-plane     & \SI{9}{\centi\meter} & 1.171 & 2.504 & 4.922 \\
        confined     & \SI{12}{\centi\meter} & 0.987 & 2.402 & \\
        \hline
        \hline
        trakSTAR         & & & & \\
        Stated Accuracy  & & \SI{1.400} & & \\
        (Lab conditions) & & & & \\
    \end{tabular}
    \caption{Comparison of tracking error (RMSE \& standard deviation) and prediction uncertainty $\sigma_{pred}$ for different online architectures in evaluation setups from \tabref{tab:scenarios}.\\
    $^1$ Linear regression only: $\sigma_{pred} = 0.673\:mm$}
    \label{tab:results}
\end{table}


\section{Discussion}
\begin{figure}[h]
    \centering
    \includegraphics[width=1.0\linewidth]{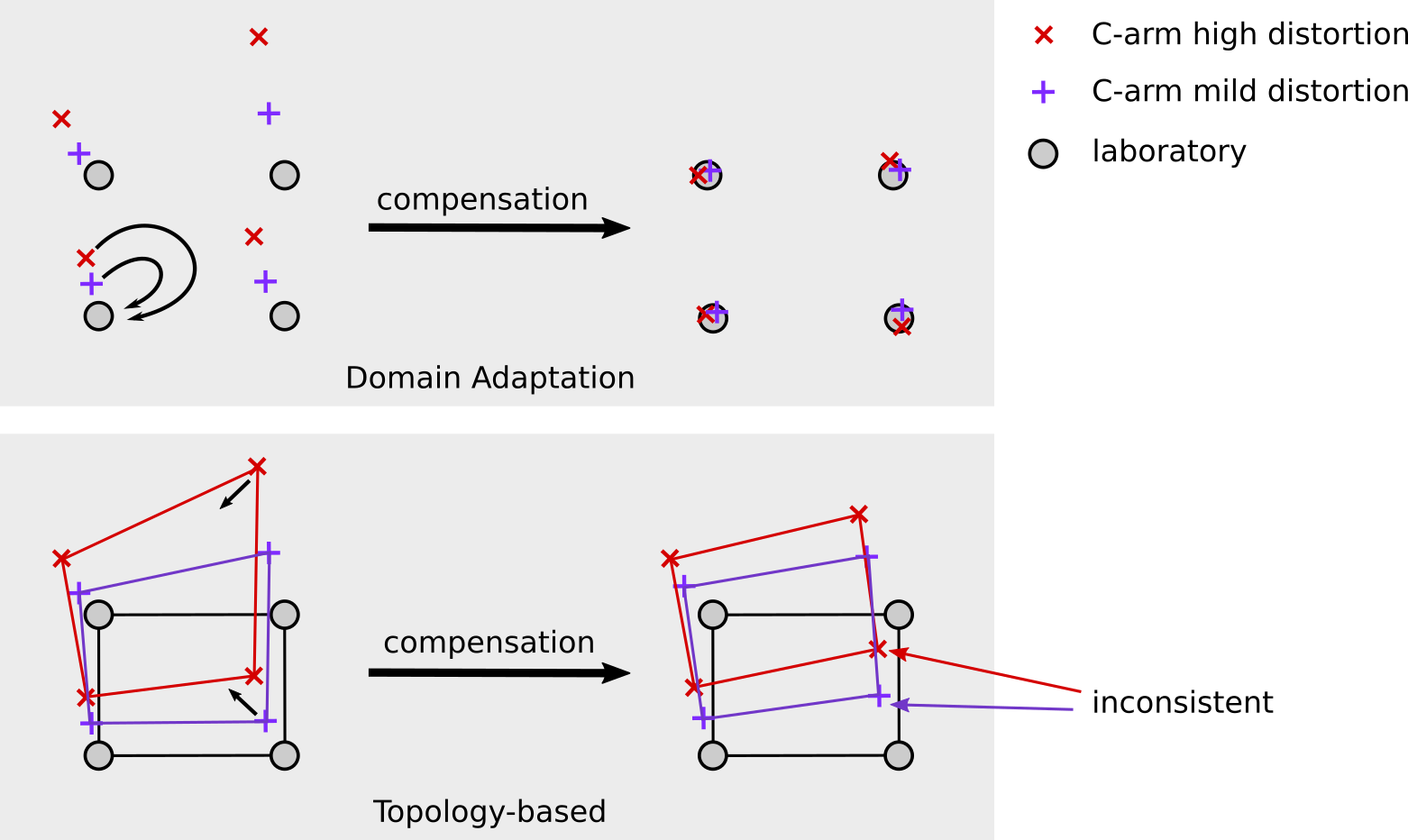}
    \caption{Schematic comparison of domain adaptation (top) versus topological (bottom) compensation schemes. }
    \label{fig:info_compensation}
\end{figure}

Comparing our domain adaptation approach to previously proposed topology-based compensation \cite{krumb2020leveraging}, we see that online compensation performance is not only a matter of RMSE, but needs to be assessed with interpretability in mind.
Predictions made by the topology-based method are hardly explicable (as seen in \figref{fig:eval_ffnn}).
We can only hypothesize that the \ac{ANN} tries to fulfill topological constraints to minimize distance error, without developing an understanding of what makes a plausible compensated point.
Figure \ref{fig:info_compensation} illustrates how domain adaptation is explicable and consistent by design, whereas topological compensation is not.

We train a modified CycleGAN on positions from various environments, achieving a translation of distorted (C-arm) to undistorted (laboratory) points.
The CycleGAN-based setup is capable of reducing error in different unseen C-arm scenarios, while still producing consistent results.
Since we can show that points are always mapped to laboratory counterparts -- regardless of these being directly present in the training set -- our approach is also explicable.
Hence, our domain adaptation approach is interpretable by design.

\section{Conclusion \& Future Work}
In this paper, we present an approach for online positional error compensation for EMT, which focuses on interpretability.
Interpretable online error compensation raises trust in \ac{EMT}, making it suitable for hybrid navigation, where the surgeon has to rely on \ac{EMT} until the next X-ray image is taken.
With reliable, online-compensated \ac{EMT} navigation, radiation exposure can be reduced to a minimum, thus bringing less harm to patient and surgeon.

Our cycle-consistent adversarial domain adaptation approach for \ac{EMT} error compensation is interpretable by design and generalizes to unseen scenarios, as we demonstrate in a prototype hybrid scenario.
As our method was originally designed to only compensate positional error, it is surprising to see that it has potential to correct rotational inaccuracies as well.

Consequently, we plan to investigate rotational compensation in the future.
Once our approach yields verifiable results on the rotational axes, it is ready for further bedside evaluations incorporating feedback of surgeons who use the hybrid system.
Finally, more evaluations on the bedside bring us closer to make hybrid \ac{EMT} navigation clinical reality.

\begin{acknowledgements}
We would like to thank Prof. Dr. med. Wolfram Voelker at Universitätsklinikum Würzburg for providing us the aortic phantom, which added significant value to our paper.
\end{acknowledgements}

\section*{Compliance with Ethical Standards}\label{CES}

\noindent\textbf{Disclosure of potential conflicts of Interest: }
This research was partially funded by the German Research Foundation.
The authors declare that they have no conflict of interest.

\noindent\textbf{Research involving Human Participants and/or Animals: }
This article does not contain any studies with human participants or animals performed by any of the authors.

\noindent\textbf{Informed consent: }
This articles does not contain patient data.

%
%
%
\bibliographystyle{splncs04}
\bibliography{00_paper}

\begin{thebibliography}{10}
\providecommand{\url}[1]{\texttt{#1}}
\providecommand{\urlprefix}{URL }
\providecommand{\doi}[1]{https://doi.org/#1}

\bibitem{aoki2020laparoscopic}
Aoki, T., Mansour, D.A., Koizumi, T., Wada, Y., Enami, Y., Fujimori, A.,
  Kusano, T., Matsuda, K., Nogaki, K., Tashiro, Y., et~al.: Laparoscopic liver
  surgery guided by virtual real-time ct-guided volume navigation. Journal of
  Gastrointestinal Surgery pp.~1--8 (2020)

\bibitem{dijkstra2011intraoperative}
Dijkstra, M.L., Eagleton, M.J., Greenberg, R.K., Mastracci, T., Hernandez, A.:
  Intraoperative c-arm cone-beam computed tomography in fenestrated/branched
  aortic endografting. Journal of vascular surgery  \textbf{53}(3),  583--590
  (2011)

\bibitem{franz2014electromagnetic}
Franz, A.M., Haidegger, T., Birkfellner, W., Cleary, K., Peters, T.M.,
  Maier-Hein, L.: Electromagnetic tracking in medicine—a review of
  technology, validation, and applications. IEEE transactions on medical
  imaging  \textbf{33}(8),  1702--1725 (2014)

\bibitem{goodfellow2014generative}
Goodfellow, I., Pouget-Abadie, J., Mirza, M., Xu, B., Warde-Farley, D., Ozair,
  S., Courville, A., Bengio, Y.: Generative adversarial nets. In: NEURIPS. pp.
  2672--2680 (2014)

\bibitem{kazeminia2020GANs}
Kazeminia, S., Baur, C., Kuijper, A., {van Ginneken}, B., Navab, N.,
  Albarqouni, S., Mukhopadhyay, A.: Gans for medical image analysis. Artificial
  Intelligence in Medicine  \textbf{109},  101938 (2020).
  \doi{https://doi.org/10.1016/j.artmed.2020.101938},
  \url{http://www.sciencedirect.com/science/article/pii/S0933365719311510}

\bibitem{kindratenko2000survey}
Kindratenko, V.V.: A survey of electromagnetic position tracker calibration
  techniques. Virtual Reality  \textbf{5}(3),  169--182 (2000)

\bibitem{kindratenko2005neural}
Kindratenko, V.V., Sherman, W.R.: Neural network-based calibration of
  electromagnetic tracking systems. Virtual Reality  \textbf{9}(1),  70--78
  (2005)

\bibitem{kingma2014adam}
Kingma, D.P., Ba, J.: Adam: A method for stochastic optimization. arXiv
  preprint arXiv:1412.6980  (2014)

\bibitem{krumb2020leveraging}
Krumb, H., Hofmann, S., K{\"u}gler, D., Ghazy, A., Dorweiler, B., Bredemann,
  J., Schmitt, R., Sakas, G., Mukhopadhyay, A.: Leveraging spatial uncertainty
  for online error compensation in emt. IJCARS pp.~1--9 (2020)

\bibitem{kuegler2019high}
K{\"u}gler, D., Krumb, H., Bredemann, J., Stenin, I., Kristin, J., Klenzner,
  T., Schipper, J., Schmitt, R., Sakas, G., Mukhopadhyay, A.: High-precision
  evaluation of electromagnetic tracking. IJCARS  \textbf{14}(7),  1127--1135
  (2019)

\bibitem{lakshminarayanan2017simple}
Lakshminarayanan, B., Pritzel, A., Blundell, C.: Simple and scalable predictive
  uncertainty estimation using deep ensembles. In: Advances in neural
  information processing systems. pp. 6402--6413 (2017)

\bibitem{nash1951non}
Nash, J.: Non-cooperative games. Annals of mathematics pp. 286--295 (1951)

\bibitem{pytorch}
Paszke, A., Gross, S., Massa, F., Lerer, A., Bradbury, J., Chanan, G., Killeen,
  T., Lin, Z., Gimelshein, N., Antiga, L., Desmaison, A., Kopf, A., Yang, E.,
  DeVito, Z., Raison, M., Tejani, A., Chilamkurthy, S., Steiner, B., Fang, L.,
  Bai, J., Chintala, S.: Pytorch: An imperative style, high-performance deep
  learning library. In: Wallach, H., Larochelle, H., Beygelzimer, A.,
  d'Alch\'{e} Buc, F., Fox, E., Garnett, R. (eds.) NEURIPS 32, pp. 8024--8035.
  Curran Associates, Inc. (2019)

\bibitem{sadjadi2016simultaneous}
{Sadjadi}, H., {Hashtrudi-Zaad}, K., {Fichtinger}, G.: Simultaneous
  electromagnetic tracking and calibration for dynamic field distortion
  compensation. TBME  \textbf{63}(8),  1771--1781 (2016)

\bibitem{szegedy2016rethinking}
Szegedy, C., Vanhoucke, V., Ioffe, S., Shlens, J., Wojna, Z.: Rethinking the
  inception architecture for computer vision. In: Proceedings of the IEEE
  conference on computer vision and pattern recognition. pp. 2818--2826 (2016)

\bibitem{tjoa2020survey}
Tjoa, E., Guan, C.: A survey on explainable artificial intelligence (xai):
  Towards medical xai. IEEE Trans. Neural Netw. Learn. Syst  \textbf{PP} (2020)

\bibitem{zhu2017unpaired}
Zhu, J.Y., Park, T., Isola, P., Efros, A.A.: Unpaired image-to-image
  translation using cycle-consistent adversarial networks. In: ICCV (2017)

\end{thebibliography}

\end{document}